\documentclass[10pt,twocolumn,letterpaper]{article}

\usepackage[pagenumbers]{cvpr} 

\usepackage{graphicx}
\usepackage{amsmath}
\usepackage{amssymb}
\usepackage{booktabs}
\usepackage{multirow}
\usepackage{color}

\newcommand{\blue}[1]{\textcolor{blue}{#1}}

\newcommand{\green}[1]{\textcolor{green}{#1}}

\usepackage[pagebackref=false,breaklinks=true,letterpaper=true,colorlinks,urlcolor=blue,citecolor=blue,linkcolor=blue,bookmarks=false]{hyperref}

\usepackage[capitalize]{cleveref}
\crefname{section}{Sec.}{Secs.}
\Crefname{section}{Section}{Sections}
\Crefname{table}{Table}{Tables}
\crefname{table}{Tab.}{Tabs.}

\def\cvprPaperID{****} 
\def\confName{CVPR}
\def\confYear{2023}

\begin{document}

\title{Pillar R-CNN for Point Cloud 3D Object Detection}

\author{Guangsheng Shi\\
Harbin Institute of Technology\\
{\tt\small sgsadvance@163.com}
\and
Ruifeng Li $\star$\\
Harbin Institute of Technology\\
{\tt\small lrf100@hit.edu.cn}
\and
Chao Ma $\star$\\
Shanghai Jiao Tong University\\
{\tt\small chaoma@sjtu.edu.cn}
}
\maketitle

\begin{abstract}
The performance of point cloud 3D object detection hinges on effectively representing raw points, grid-based voxels or pillars. 
Recent two-stage 3D detectors typically take the point-voxel-based R-CNN paradigm, i.e., the first stage resorts to the 3D voxel-based backbone for 3D proposal generation on bird-eye-view (BEV) representation and the second stage refines them via the intermediate point representation.
Their primary mechanisms involve the utilization of intermediary keypoints to restore the substantial 3D structure context from the converted BEV representation.
The skilled point-voxel feature interaction, however, makes the entire detection pipeline more complex and compute-intensive.
In this paper, we take a different viewpoint -- the pillar-based BEV representation owns sufficient capacity to preserve the 3D structure.
In light of the latest advances in BEV-based perception, we devise a conceptually simple yet effective two-stage 3D detection architecture, named Pillar R-CNN.
On top of densified BEV feature maps, Pillar R-CNN can easily introduce the feature pyramid architecture to generate 3D proposals at various scales and take the simple 2D R-CNN style detect head for box refinement.
Our Pillar R-CNN performs favorably against state-of-the-art 3D detectors on the large-scale Waymo Open Dataset but at a small extra cost.
It should be highlighted that further exploration into BEV perception for applications involving autonomous driving is now possible thanks to the effective and elegant Pillar R-CNN architecture.
\end{abstract}

\section{Introduction}
\label{sec:intro}

Point cloud 3D object detection plays a crucial role in 3D scene understanding for robotics and autonomous driving. 
However, compared with well-developed image-based 2D detection, 
LiDAR-based 3D detection still struggles to cope with the sparse and irregular nature of point clouds.
In this paper, we propose Pillar R-CNN, a Faster R-CNN-like architecture on pillar-based point cloud representation that can profit from the advances of the 2D detection field.

The performance of LiDAR-based 3D detection hinges on representation learning on point clouds.
Deep convolutional feature backbones include the point-based \cite{qi2017pointnet++} or grid-based (\textit{e.g.}, 3D voxel-based \cite{zhou2018voxelnet,yan2018second} and 2D pillar-based \cite{lang2019pointpillars}) representations.
Current state-of-the-art approaches \cite{shi2020pv,shi2020points,deng2020voxelrcnn,mao2021pyramid} mainly apply one-stage detectors to produce class-specific proposals and require abstracted point-wise features to make further box refinement.
Typically, PV-RCNN \cite{shi2020pv} extends SECOND \cite{yan2018second} by introducing the precisely positioned keypoints from raw points to preserve the significant 3D structure information. Voxel Set Abstraction (VSA) is introduced to enrich each keypoints with the multi-scale 3D voxel feature context.
Each 3D RoI feature is further extracted from keypoints through RoI-grid pooling for box refinement.
Despite the decent detection accuracy, the unordered storage of keypoints leads to costly computation overheads.
Voxel R-CNN \cite{deng2020voxelrcnn} argues that coarse 3D voxels can also offer sufficient detection accuracy. Voxel RoI Pooling extracts 3D RoI features from the 3D sparse feature volumes. The regularity of 3D voxels facilitates the search for nearby voxel features while its coarse granularity sacrifices the detection accuracy especially for small objects.
Besides, Part-\textit{A}$^2$ \cite{shi2020points} takes the upsampled 3D voxel points as transitional keypoints to alleviate the issue of overly coarse voxel granularity. 
Moreover, most recent works \cite{mao2021pyramid,hu2022point} propose the variants of the RoI-grid pooling module to handle the issues of sparsity and point density variation of intermediary keypoints.
In a nutshell, top-performing two-stage 3D detection frameworks heavily depend on the 3D voxel-based backbone for BEV-based 3D proposal generation, and then restore the 3D structure context to make further box refinement based on the point-level features.
Nevertheless, the point-voxel-based detection paradigm and its associated point-based set operations \cite{qi2017pointnet,shi2020pv,deng2020voxelrcnn} complicate the entire 3D detection pipeline and require extra efforts for boosting efficiency.

The latest progress on BEV-based perception \cite{huang2021bevdet,ma2022vision,shi2022pillarnet,mao20223d} demonstrates the potential of BEV representation for high-performance 3D object detection. 
Pillar, as a typical phenotype of BEV representation on point clouds, might supply sufficient 3D structure information and can easily integrate the advances of mature 2D detection fields.
In this work, towards this objective, we take solely on pillar-based point cloud representation and try to boost its accuracy.
We first argue that pillars with proper granularity can also offer the significant 3D structure for box refinement.
As such, we propose a surprisingly flexible and effective two-stage framework, named \textit{Pillar R-CNN}, that integrates FPN \cite{lin2017feature} with region proposal network (RPN) for class-specific 3D proposal generation at various scales and then crop the 2D dense pooling feature maps for further box refinement at a single manageable scale.
Specifically, we use a pillar-based feature backbone with a hybrid of sparse and dense convolutions to compute hierarchical feature maps in one forward pass.
To counteract the degraded accuracy on small objects, we devise a lateral connection layer to build a pillar-based pyramidal backbone like FPN \cite{lin2017feature} by densifying sparse pillar volumes. 
However, this is too difficult and complex to be utilized on the 3D voxel-based feature backbone \cite{yan2018second}.
\cref{tab:ab_any} shows that incorporating FPN into RPN produces better class-specific proposals for small objects, just like its 2D counterpart \cite{he2017mask}.
Moreover, we follow the RoIAlign \cite{ren2015faster} to crop the dense pooling maps to refine 3D proposals. Here, the pooling maps are built by another lateral connection layer over the feature hierarchies with manageable resolution.
Besides, the single-scale pooling maps are class-agnostic and robust to object scale variation.
This slightly differs from the commonly used FPN \cite{lin2017feature} on Faster R-CNN \cite{ren2015faster}.

Despite refining 3D proposals on a $4 \times$ downsampled 2D feature map (\textit{i.e.}, pillar size of 0.4m), as shown in \cref{tab:waymo_val}, the proposed Pillar R-CNN achieves comparable detection accuracy with previous state-of-the-art two-stage methods on the large-scale Waymo Open Dataset \cite{sun2020scalability}.

The main contribution of this work stems from the conceptually simple yet effective design of Pillar R-CNN, which demonstrates the powerful modeling capability of pillar-based BEV representation on point clouds.
The impressive experiment results of Pillar R-CNN also confirm our viewpoint: intermediate point-level representation in a two-stage approach is not crucial for high-performance 3D object detection and BEV representation with proper pillar granularity can also afford sufficient 3D structure for this task.
Without bells and whistles, Pillar R-CNN bridges the domain gap between LiDAR-based 3D detection and image-based 2D detection by means of pillar-based BEV representation.
Despite only demonstrating BEV representation with Pillar R-CNN on point clouds, we believe the findings presented are equally applicable to the BEV-based detectors.


\section{Related Work}

\paragraph{Two-stage 2D image object detection.}
Representative two-stage 2D object detection methods \cite{girshick2015fast,he2017mask,ren2015faster} take a conceptually simple yet effective paradigm.
The first stage, called the region proposal network (RPN), generates an axis-aligned Region of Interest (RoI) by sliding windows over all locations in a class-agnostic manner.
The second stage crops the RoI feature by RoIPool \cite{girshick2015fast} or RoIAlign \cite{he2017mask}, and then conducts proposal-specific classification and bounding-box regression.
The feature pyramid network \cite{lin2017feature}, as a basic component recognition system, can build pyramidal multi-level feature maps that hold high-level semantic features at all scales.
Our proposed 3D detection method shares the same spirit of FPN and RoI pooling but utilizes domain-specific techniques to address the issues of sparse and irregular point clouds on 2D BEV space.

\vspace{-3mm}
\paragraph{Single-stage point cloud 3D object detection.}
Single-stage 3D detection methods can be mainly divided into three streams, \textit{i.e.,} point-based, voxel-based and pillar-based. 
Point-based single-stage detectors directly learn point-wise features from raw point clouds, where set abstraction \cite{qi2017pointnet,qi2017pointnet++} enables flexible receptive fields by setting different search radii.
3DSSD \cite{yang20203dssd} introduces F-FPS for point downsampling and first performs 3D detection only using an encoder network.
Point-based methods straightforwardly consume point clouds, however, insufficient learning capacity and unordered storage become the main bottleneck.
Voxel-based single-stage detectors usually first rasterize point clouds into 3D voxel grids to be processed by 3D dense \cite{zhou2018voxelnet} or sparse convolution \cite{graham20183d, yan2018second} for geometry feature learning.
Those voxel-based methods can offer superior detection performance while the computational/memory overheads grow cubically with the used voxelization resolution.
Pillar-based single-stage detectors further simplify 3D voxels to 2D pillars to be processed by 2D convolutions, paving the way for embedded deployments.
PointPillars \cite{lang2019pointpillars} first encodes the input point clouds into regular pillars and utilizes a simple top-down network for final 3D object detection.
The latest PillarNet \cite{shi2022pillarnet} deeply the huge performance gap of voxel- and pillar-based methods in terms of the architecture components and proposes the ``encoder-neck-head" pipeline for a better accuracy/speed trade-off.
Moreover, we deeply dive into the pillar-based point cloud representation for better detection accuracy on PillarNet \cite{shi2022pillarnet}.

\vspace{-3mm}
\paragraph{Two-stage point cloud 3D object detection.}
Two-stage 3D detection methods typically build on top of single-stage methods for generating class-specific 3D proposals, but focus on the proposal-specific feature extraction for box refinement based on costly point set abstraction operation.
PointRCNN \cite{shi2019pointrcnn} first applies PointNet++ \cite{qi2017pointnet} as the feature backbone for bottom-up 3D proposal generation and proposes a novel point cloud region pooling for 3D box refinement.
Similarly, STD \cite{yang2019std} generates point-based proposals from raw point clouds, but proposes a new PointsPool layer to introduce volumetric representation for compact RoI feature extraction.
Fast Point R-CNN \cite{ronneberger2015u} introduces the voxel-based backbone for producing high-quality 3D proposals and applies an attention-based PointNet pooling module for box refinement. 
PV-RCNN \cite{shi2020pv} firstly integrates the point- and voxel-based feature learning schemes, where a small set of raw points as intermediate keypoints are enriched with multi-scale 3D voxel features and the 3D RoI features are extracted through RoI-grid pooling for box refinement.
Incrementally, Voxel R-CNN \cite{deng2020voxelrcnn} substitutes the precise keypoints with coarse voxel center points with better memory locality to facilitate the search for nearby voxel features. 
Part-\textit{A}$^2$ \cite{shi2020points} utilizes a UNet-like \cite{ronneberger2015u} backbone for both 3D proposal generation and intra-object part location prediction, and then designs a novel RoI-aware point cloud pooling module for exact box refinement without ambiguity.
The most recent Pyramid R-CNN \cite{mao2021pyramid} and PDV \cite{hu2022point} design the variants of RoI-grid pooling to resolve the sparsity and non-uniform distribution of point clouds for better detection accuracy.
Top-performing point-voxel-based 3D detectors rely on the voxel-based backbone for BEV-based 3D proposal generation but resort to intermediate point-level representation for further refinement.
Differently, we convert point clouds to regular pillars and conduct 3D proposal generation and box refinement soly on the BEV representation without using intermediate keypoints.

\begin{figure*}[th]
  \centering
  \begin{subfigure}{0.495\linewidth}
  \centering
    \includegraphics{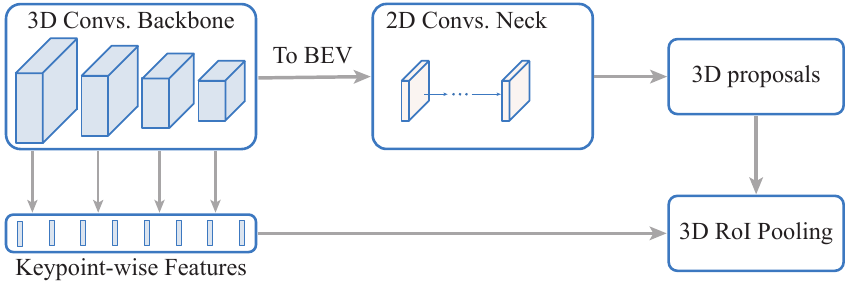}
    \caption{Point-Voxel-based 3D R-CNN}
  \end{subfigure}
  \hfill
  \begin{subfigure}{0.495\linewidth}
    \centering
    \includegraphics{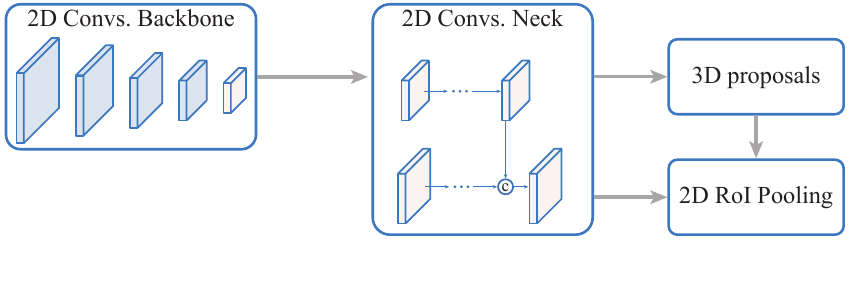}
    \caption{Our Pillar R-CNN}
  \end{subfigure}
  \caption{Compared with prior best-performing point-voxel-based 3D detection paradigms, our proposed Pillar R-CNN directly refines 3D RoI on the 2D BEV space as Faster R-CNN without the necessity of intermediary keypoint-wise features.}
  \label{fig:comparison}
\end{figure*}

\section{Preliminary}

In this section, we first revisit best-performing two-stage 3D detectors and then reflect on the performance bottleneck of point cloud 3D detection referring to image-based 2D detection. 

\subsection{Revisiting}
Current state-of-the-art two-stage 3D object detection approaches \cite{shi2020pv,deng2020voxelrcnn,shi2020points}, as in \cref{fig:comparison}, typically employ the point-voxel-based feature learning scheme for the BEV-based 3D proposal generation and then conduct point-level box refinement in the 3D space. 
In both of these two stages, effective keypoint representation plays the primary role.
The pioneering PV-RCNN \cite{shi2020pv} sub-samples raw point clouds via Farthest Point Sampling (FPS) as intermediate keypoints and take point-based set operations \cite{shi2020pv,qi2017pointnet} for point-voxel interaction. 
Despite the impressive detection accuracy, PV-RCNN suffers from the time-consuming procedures of point sampling and neighbor searching. 
Incrementally, Voxel R-CNN \cite{deng2020voxelrcnn} argues that the coarse 3D voxels instead of precise positioning of raw points are sufficient for accurate localization on large objects.
Voxel R-CNN treats sparse but regular 3D volumes as a set of non-empty voxel center points and utilizes an accelerated PointNet module for a new balance between accuracy and efficiency.
Part-\textit{A}$^2$ \cite{shi2020points} adopts a UNet-like architecture for finer voxel points and gains further profits for small objects.
Most recent works \cite{mao2021pyramid,hu2022point} attempt to tackle the issues of sparsity and point density variation of intermediary keypoints for improving detection accuracy.
In summary, typical point-voxel-based 3D detectors convert point clouds to regular grids for BEV-based 3D proposal generation and hinge on the granularity of keypoints for further box refinement. This inevitably exacerbates the complexity of the detection system and also struggles with more efficient local aggregation operations like \cite{deng2020voxelrcnn}.

From the perspective of BEV-based perception, we expect to design a compact R-CNN head on fine pillars for box refinement.
By taking a close at the 3D voxel-based detectors \cite{yan2018second,Hu2021AFDetV2RT}, we observe the fact that they convert 3D feature volumes into BEV representations for the dense detect head.
We conjecture that the merely bird-eye-view (BEV) representation can still offer sufficient 3D structure information. 
To confirm our viewpoint, we construct the reasonable pooling feature map with a manageable granularity and crop this 2D dense map for each 3D proposal for 3D RoI refinement. 
Also, the encouraging results convince us that BEV representation can carry sufficient 3D structure information without the need to restore this context via point-level representation.

\subsection{Domain Gap between Two-Stage 2D and 3D Detection}

Unlike the well-studied 2D community where the input image is a 2D densely packed array, the inherent sparsity and irregularity of point clouds make it divert from the 2D detection field in methodology.
The main gap may be the 3D point/voxel-based point cloud representation and transitional keypoints in the two-stage frameworks. 
Due to this gap, the latest progress in 2D detection cannot be easily applied to LiDAR-based 3D detection. 
For example, the feature pyramid network (FPN) \cite{lin2017feature}, as a basic component in 2D detection, has not yet been successfully used by current LiDAR-based 3D detection. 
The latest advance in pillar-based 3D detection \cite{shi2022pillarnet} partially bridges this gap by introducing the well-developed backbones, such as VGGNet \cite{simonyan2014very} and ResNet \cite{he2016deep}, in 2D detection to 3D object detection with demonstrated success.
Along this research direction, we aim to narrow the domain gap by developing a 2D R-CNN style detector as \cite{ren2015faster} for Lidar-based 3D detection. 
The proposed Pillar R-CNN integrates the basic FPN with RPN properly for small object detection and crops the 2D dense pooling feature map like RoIPool \cite{chen2019fast} or RoIAlign \cite{ren2015faster} for further box refinement without the usage of transitional keypoint representation.

\begin{figure*}[t]
\centering
\includegraphics[]{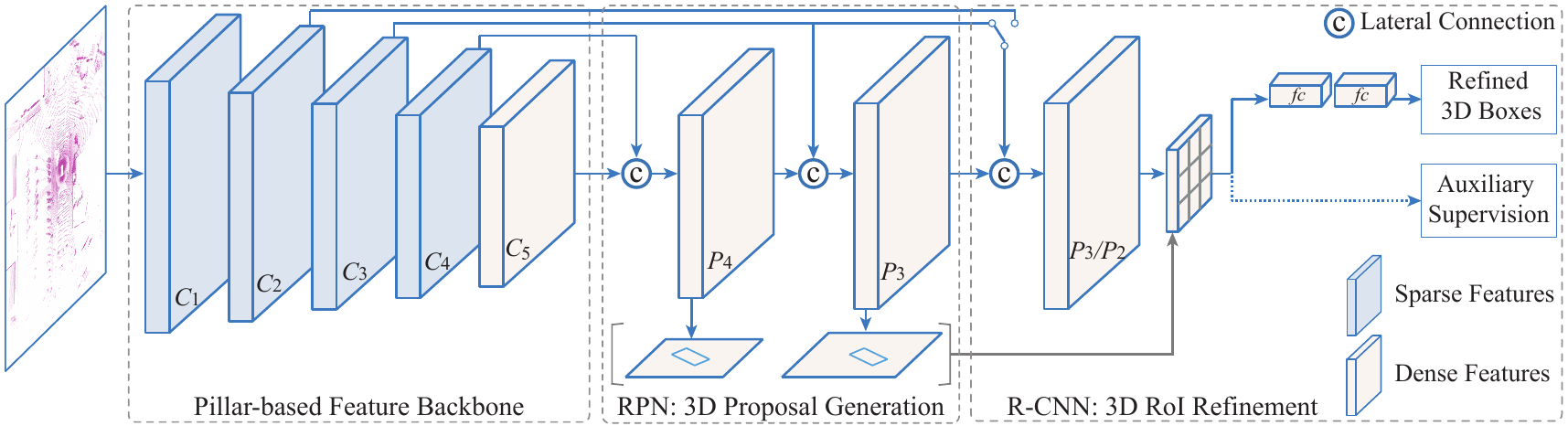}
\caption{Overall architecture of our proposed Pillar R-CNN.
The pillar-based backbone establishes a feature hierarchy in one forward pass, resulting in multi-level feature maps.
The in-network feature pyramid of RPN creates feature maps at different spatial resolutions that have rich semantics at all class-related scales.
RPN attaches the same detect head to each pyramid level to take class-specific detection.
R-CNN refines 3D RoI proposals at a manageable resolution by creating class-agnostic 2D dense pooling feature maps via the novel lateral connection.
The dashed arrow indicates the detachable auxiliary segmentation branch.}
\label{fig:framework}
\end{figure*}

\section{Pillar R-CNN}

As shown in \cref{fig:framework}, our proposed Pillar R-CNN is conceptually simple, including two stages: the first stage (RPN) produces 3D proposals along with its classification at all class-related scales; the second stage (R-CNN) refines 3D boxes on the BEV plane at a manageable resolution.

\subsection{Pillar-based Backbone}
The most recent pillar-based 3D detector, PillarNet \cite{shi2022pillarnet}, converts point clouds to regular pillars and handles the sparse 2D volumes hierarchically in a bottom-up pathway. 
PillarNet takes the advantage of 2D object detection with ConvNets such as VGGNet \cite{simonyan2014very} and ResNet \cite{he2016deep} and computes the feature hierarchy in one forward pass.
It creates a set of low-level sparse 2D pillar volumes and high-level dense feature maps via the hybrid of 2D sparse and dense convolutions in a bottom-up manner.
We denote the resulting multi-level feature volumes or maps from different pyramid levels by $\{C_1, C_2, C_3, C_4, C_5\}$ with strides $\{1, 2, 4, 8, 16\}$ of pillar scales.

\subsection{Pyramidal Region Proposal Network}
Similar to the basic FPN \cite{lin2017feature}, we construct a pyramidal region proposal network to improve the detection accuracy of small objects like pedestrians.
To build high-level semantic feature maps at all class-related scales, we modify the lateral connection layer to effectively merge the top-down dense maps and bottom-up sparse volumes.
Each lateral connection merges the sparse feature volumes and dense feature maps of the same spatial size from the bottom-up pathway and the top-down pathway.
Specifically, the top-down pathway hallucinates higher resolution features by upsampling semantically stronger dense feature maps from higher pyramid levels with a stride 2 de-convolution layer.
The upsampled dense map is then merged with the corresponding densified bottom-up map from its sparse volumes by the simple concatenation.
This process is iterated until the required multi-scale feature maps are ready.
Finally, we append a $3 \times 3$ convolutional layer on each merged map to alleviate the aliasing effect of upsampling and reduce the channel dimensions.
Here, the final set of feature maps with high-level rich semantics called $\{P_3, P_4\}$, correspond to $\{C_3, C_4\}$ respectively in terms of the spatial size.

We attach the commonly used center detect head \cite{yin2021center} to each level of in-network feature pyramid $\{P_3, P_4\}$, on which class-specific object predictions are respectively made.
Although better lateral connection designs are more helpful, we stick to the simplest design as described above.

\subsection{R-CNN via 2D RoI Pooling on BEV Plane}
To justify the BEV representation preserves the crucial 3D structure information, we construct the dense pooling feature maps at a manageable scale via our plainly designed lateral connection layer.
Then, we make further 3D box refinement by using a Faster R-CNN-like RoI pooling module on the BEV plane.
We additionally supervise the grid points of each 3D proposal with semantical supervision to further examine the 3D structural capability of BEV representation, aligning the extra keypoint segmentation supervision branch used in \cite{shi2020pv}.

\begin{figure}[h]
\centering
\includegraphics[width=0.5\linewidth]{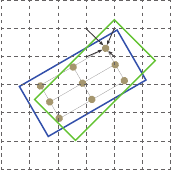}
\caption{Pooling points (brown) feature extraction on the pooling map (dashed grid) from nearby grid points using bilinear interpolation. 
\blue{Blue box}: 3D RoI. \green{Green box}: ground truth.}
\label{fig:interpolation}
\end{figure}

\vspace{-3mm}
\paragraph{Lateral connection layer.}
The aim of the lateral connection layer for building pooling maps is to integrate the low-level sparse pillar feature volumes and high-level dense semantic feature maps into the dense pooling maps at a manageable resolution.
Specifically, the 2D sparse convolution layer is used to downsample the selected low-level pillar volumes from the bottom-up pathway to the desired spatial resolution.
Then, the stride $2$ de-convolution layer upsamples the dense semantic feature maps from the pyramidal neck module in a top-down manner.
Finally, the dense pooling maps are constructed by merging the semantically stronger dense maps and spatially more precise sparse volumes with the same spatial resolution via a $3 \times 3$ convolutional layer.
Our designed lateral connection layer slightly deviates from the element-wise addition way in the original FPN \cite{lin2017feature}, in that the sparse volumes could be extremely sparse with mostly empty points.
To confront the issue of the center feature missing induced by the sparse features, we simply apply the concatenation and standard convolution operations to blend the densified maps from the bottom-up pathway and dense maps from the top-down pathway.
In this way, our lateral connection design can provide critical semantic coverage onto the spatially accurate yet sparse bottom-up pillar volumes.

\vspace{-3mm}
\paragraph{Auxiliary segmentation supervision.}
We attach the detachable auxiliary segmentation supervision of grid points in each 3D proposal to promote the 3D structural capability of the BEV-based pooling maps, which is inspired by the extra segmentation supervision of keypoint with 3D voxel-based structure context in \cite{shi2020pv}.
 We attach the detachable auxiliary segmentation supervision of grid points in each 3D proposal to promote the 3D structural capability of the BEV-based pooling maps, which is inspired by the extra segmentation supervision of keypoint with 3D voxel-based structure context in \cite{shi2020pv}.
Specifically, we employ a 2-layer MLP with the sigmoid function to predict the foreground/background score of each grid point in each projected 2D rotated RoI. 
The segmentation layers can be directly produced from the 3D box annotations by identifying whether each grid point is inside or outside of a projected ground-truth rotated box in \cref{fig:interpolation}, since the 3D objects do not overlap on the BEV plane.
Additionally, the auxiliary segmentation supervision branch is only employed during training, so there is no additional computational overhead for inference.

Similar to the RoI pooling \cite{ren2015faster} and RoI Align \cite{he2017mask}, we crop the dense pooling maps using the projected 2D rotated RoI in order to extract the 3D RoI feature on the BEV plane.
Specifically, we simply utilize bilinear interpolation operation to sample evenly distributed $G \times G$ grid points in each projected 3D RoI in \cref{fig:interpolation}.
Then we collapse the sampled grid point features $\mathcal{R}^{G \times G \times C}$ per RoI into a vector $\mathcal{R}^{256}$ by two hidden 256-\textit{D} fully-connected (\textit{fc}) layers, as the common practice in previous two-stage 3D detection methods.
As a result, Our Faster R-CNN-like 2D RoI pooling module transforms each 3D RoI feature with objects of various sizes into a fixed spatial extent for 3D proposal refinement and confidence prediction.
It is worth noting that our second R-CNN stage slightly differs from Faster R-CNN \cite{ren2015faster} in that it requires the pooling maps at the manageable scale with the amply fused semantic-spatial features via the lateral connection layer.

\subsection{Training Losses}

The proposed Pillar R-CNN framework is trained end-to-end with the region proposal loss $\mathcal{L}_{rpn}$, the proposal refinement loss $\mathcal{L}_{rcnn}$, and the auxiliary segmentation loss $\mathcal{L}_{seg}$.
We adopt the same region proposal loss in \cite{yin2021center} for used pyramid scales.
We follow \cite{shi2020pv} and use the same proposal refinement loss for class-agnostic confidence prediction and box regression. 
The auxiliary segmentation is simply supervised using a binary cross-entropy loss, with each grid point's classification label determined by its projected location relative to the corresponding ground-truth bounding box.
The overall training loss is the sum of these three losses with equal weights:
\begin{equation}
\centering
\mathcal{L}_{total} = \sum_{s}{\mathcal{L}_{rpn}^s} + \mathcal{L}_{rcnn} + \mathcal{L}_{seg}
\label{eq:total-loss}
\end{equation}

\section{Experiments}

We evaluate our Pillar R-CNN on the public Waymo Open Dataset \cite{sun2020scalability}, a large-scale 3D object detection dataset for autonomous driving research.
The whole dataset is composed of 798 training sequences (158,361 point cloud samples) and 202 validation sequences (40,077 point cloud samples).
The evaluation metrics used are 3D mean Average Precision (mAP) and mAP weighted by heading accuracy (mAPH).
The mAP and mAPH are based on an IoU threshold of 0.7 for vehicle and 0.5 for pedestrian and cyclist.
Two difficulty levels, LEVEL\_1 (boxes with more than five LiDAR points) LEVEL\_2 APH (boxes with at least one LiDAR point) are considered.
We highlight LEVEL\_2 APH in tables since it is the main metric for ranking in the Waymo Challenge Leaderboard.

\vspace{-3mm}
\paragraph{Implementation details.}
The architecture of the pillar-based backbone follows the design of PillarNet \cite{shi2022pillarnet}, which includes the first four stages with sparse convolution for processing sparse pillar volumes and the last stage with standard convolution for densified feature maps on the BEV plane.
Based on the inherent feature hierarchy of pillar-based backbone, we simply introduce FPN into RPN by detecting large object detection (\textit{e.g.}, vehicle) on $8\times$ stridden maps and small objects (\textit{e.g.}, pedestrian) on $4\times$ stridden maps.
The second stage constructs the pooling maps at a manageable resolution and channel dimension via a $3 \times 3$ convolutional layer. We only conduct experiments on the pooling resolutions of $P_2$ and $P_3$, but not using $P_1$ due to its large memory footprint.
Unless mentioned, we employ the $4\times$ stridden pooling maps of the pillar size 0.4m, because it offers a good trade-off between localization performance and computation costs.

Following the settings of the baseline \cite{shi2022pillarnet}, the detection range is [-75.2m, 75.2m] for the X and Y axis, and [-2m, 4m] for the Z axis. We also set the initial pillar size to (0.1m, 0.1m).

\vspace{-3mm}
\paragraph{Training and inference details.}
Following CenterPoint \cite{yin2021center}, our Pillar R-CNN framework is trained from scratch in an end-to-end manner with the AdamW optimizer, batch size 16 and learning rate 0.01 for 36 epochs on 4 RTX 3090 GPUs on the Waymo Open Dataset.
We report inference latency on a single 3090 GPU with batch size 1 following the way in \cite{yin2021center} without any test-time optimization.
We randomly sample 128 proposals with 1:1 ratio as the training samples for the proposal refinement stage, where the positive proposals have at least 0.55 3D IoU with the corresponding ground-truth boxes.
The data augmentation strategies during training are kept the same as PV-RCNN \cite{shi2020pv} for a fair comparison.

During inference, we keep the top (200, 150, 150) proposals generated by RPN for the vehicle, pedestrian, and cyclist with class-specific NMS, where the used 3D IoU thresholds are set to be 0.8, 0.55 and 0.55 for vehicle, pedestrian and cyclist, respectively.

\begin{table*}[h]
\centering
\resizebox{\linewidth}{!}{%
\begin{tabular}{c|c|c|cc|cc|cc|cc}
\toprule
\multirow{2}{*}{Methods} & \multirow{2}{*}{Stages} & \multirow{2}{*}{Sensors} & 
\multicolumn{2}{c|}{ALL (3D AP/APH)} &
\multicolumn{2}{c|}{Veh. (3D AP/APH)} & 
\multicolumn{2}{c|}{Ped. (3D AP/APH)} & 
\multicolumn{2}{c}{Cyc. (3D AP/APH)} \\
 & & & L1 & L2 & L1 & L2 & L1 & L2 & L1 & L2 \\ 
 \midrule
StarNet \cite{ngiam2019starnet} & Two & - & - & - & 61.50/61.00 & 54.90/54.50 & 67.80/59.90 & 61.10/54.00 & - & - \\
RCD \cite{bewley2020range} & Two & - & - & - & 71.97/71.59 & 65.06/64.70 & - & - & - & - \\
Light-FMFNet \cite{murhij2021real} & One & L & 71.24/67.26 & 65.88/62.18 & 77.85/77.30 & 70.16/69.65 & 69.52/59.78 & 63.62/54.61 & 66.34/64.69 & 63.87/62.28 \\
HIKVISION\_LiDAR \cite{xu2021centeratt} & Two & L & 75.19/72.58 & 69.74/67.29 & 78.63/78.14 & 71.06/70.60 & 76.00/69.90 & 69.82/64.11 & 70.94/69.70 & 68.35/67.15 \\
CenterPoint \cite{yin2021center} & Two & LT & - & - & 80.20/79.70 & 72.20/71.80 & 78.30/72.10 & 72.20/66.40 & - & - \\
AFDetV2 \cite{Hu2021AFDetV2RT} & One & LT & 77.56/75.20 & 72.18/69.95 & 80.49/80.43 & 72.98/72.55 & 79.76/74.35 & 73.71/68.61 & \textbf{72.43/71.23} & \textbf{69.84/68.67} \\
\midrule
PillarNet-18 & One & LT & 76.60/73.62 & 71.30/68.49 & 81.85/81.40 & 74.46/74.03 & 79.97/72.68 & 73.95/67.09 & 67.98/66.80 & 65.50/64.36 \\
PillarNet-34 & One & LT & 77.46/74.69 & 72.17/69.55 & 82.47/82.03 & 75.07/74.65 & 80.82/74.13 & 74.83/68.54 & 69.08/67.91 & 66.60/65.47 \\ \midrule
Pillar R-CNN-18 & Two & LT & 77.80/74.64 & 72.56/69.55 & 81.89/81.40 & 74.52/74.07 & 82.08/74.29 & 76.23/68.84 & 69.44/68.24 & 66.92/65.75 \\
Pillar R-CNN-34 & Two & LT & \textbf{78.29/75.28} & \textbf{73.05/70.18} & \textbf{82.53/82.10} & \textbf{75.16/74.75} & \textbf{82.90/75.50} & \textbf{77.04/70.00} & 69.43/68.25 & 66.94/65.79 \\
\bottomrule
\end{tabular}
}
\caption{Single-frame LiDAR-only non-ensemble performance comparison on the Waymo Open Dataset \textit{test} set. 
``L" and ``LT" mean ``all LiDARS" and ``top-LiDAR only", respectively.
The table is sorted by ALL APH/L2 which is the official ranking metric. The proposed Pillar R-CNN models set new state-of-the-art results.
}
\label{tab:waymo_test}
\end{table*}

\begin{table*}[t]
\centering
\resizebox{\linewidth}{!}{%
\begin{tabular}{c|c|cc|cc|cc|cc}
\toprule
\multirow{2}{*}{Methods} & \multirow{2}{*}{Stages} & \multicolumn{2}{c|}{ALL (3D mAP/mAPH)} & \multicolumn{2}{c|}{Veh. (3D mAP/mAPH)} & \multicolumn{2}{c|}{Ped. (3D mAP/mAPH)} & \multicolumn{2}{c}{Cyc. (3D mAP/mAPH)} \\
 & & L1 & L2 & L1 & L2 & L1 & L2 & L1 & L2 \\ 
\midrule
StarNet \cite{ngiam2019starnet} & Two & - & - & 53.70/- & - & 66.80/- & - & - & - \\
3D-MAN \cite{yang20213d} & Multi & - & - & 69.03/68.52 & 60.16/59.71 & 71.71/67.74 & 62.58/ 59.04 & - & - \\
RCD \cite{bewley2020range} & Two & - & - & 69.59/69.16 & - & - & - & - & - \\
$\dagger$SECOND \cite{yan2018second} & One & 67.20/63.05 & 60.97/57.23 & 72.27/71.69 & 63.85/63.33 & 68.70/58.18 & 60.72/51.31 & 60.62/59.28 & 58.34/57.05 \\
$\dagger$PointPillar \cite{lang2019pointpillars} & One & 68.87/63.33 & 62.63/57.53 & 71.60/71.00 & 63.10/62.50 & 70.60/56.70 & 62.90/50.20 & 64.40/62.30 & 61.90/59.90 \\
IA-SSD \cite{zhang2022not} & One & 69.19/64.48 & 62.28/58.08 & 70.53/69.67 & 61.55/60.80 & 69.38/58.47 & 60.30/50.73 & 67.67/65.30 & 64.98/62.71 \\
LiDAR R-CNN \cite{li2021lidar} & Two & 71.10/66.20 & 64.63/60.10 & 73.50/73.00 & 64.70/64.20 & 71.20/58.70 & 63.10/51.70 & 68.60/66.90 & 66.10/64.40 \\
MVF++ \cite{qi2021offboard} & One & - & - & 74.64/- & - & 78.01/- & - & - & - \\
RSN \cite{sun2021rsn} & Two & - & - & 75.10/74.60 & 66.00/65.50 & 77.80/72.70 & 68.30/63.70 & - & - \\
Voxel R-CNN \cite{deng2020voxelrcnn} & Two & - & - & 75.59/- & 66.59/- & - & - & - & - \\
Pyramid R-CNN \cite{mao2021pyramid} & Two & - & - & 76.30/75.68 & 67.23/66.68 & -&- &- &- \\
CenterPoint \cite{yin2021center} & Two & - & - & 76.70/76.20 & 68.80/68.30 & 79.00/72.90 & 71.00/65.30 & - & - \\
PV-RCNN \cite{shi2020pv}& Two & 73.44/69.63 & 66.80/63.33 & 77.51/76.89 & 68.98/68.41 & 75.01/65.65 & 66.04/57.61 & 67.81/66.35 & 65.39/63.98 \\
Part-A$^2$ \cite{shi2020points} & Two & 73.63/70.25 & 66.93/63.85 & 77.05/76.51 & 68.47/67.97 & 75.24/66.87 & 66.18/58.62 & 68.60/67.36 & 66.13/64.93 \\
PDV \cite{hu2022point} & Two & 73.25/69.95 & 67.21/64.15 & 76.85/76.33 & 69.30/68.81 & 74.19/65.96 & 65.85/58.28 & 68.71/67.55 & 66.49/65.36 \\
AFDetV2 \cite{Hu2021AFDetV2RT} & One & 77.18/74.83 & 70.97/68.77 & 77.64/77.14 & 69.68//69.22 & 80.19/74.62 & 72.16/66.95 & \textbf{73.72/72.74} & \textbf{71.06/70.12} \\
CenterFormer \cite{zhou2022centerformer} & One & 75.37/73.00 & 71.2/68.93 & 75.2/74.7 & 70.2/69.7 & 78.6/73.0 & 73.6/68.3 & 72.3/71.3 & 69.8/68.8 \\
\midrule
PillarNet-18 \cite{shi2022pillarnet} & One & 76.15/73.20 & 69.91/67.17 & 78.24/77.73 & 70.40/69.92 & 79.80/72.59 & 71.57/64.90 & 70.40/69.29 & 67.75/66.68 \\
PillarNet-34 \cite{shi2022pillarnet} & One & 77.32/74.60 & 70.97/68.43 & 79.09/78.59 & 70.92/70.46 & 80.59/74.01 & 72.28/66.17 & {72.29/71.21} & {69.72/68.67} \\ \midrule
Pillar R-CNN-18 & Two & 77.31/74.11 & 71.27/68.24 & 78.70/78.19 & 70.53/70.06 & 82.15/74.24 & 74.78/67.27 & 71.07/69.91 & 68.50/67.38  \\
Pillar R-CNN-34 & Two & \textbf{78.12/75.02} & \textbf{72.07/69.11} & \textbf{79.47/78.98} & \textbf{71.29/70.84} & \textbf{82.67/75.06} & \textbf{75.29/68.35} & 72.21/71.01 & 69.62/68.45 \\ \midrule
Pillar R-CNN-18 (6 epoch) & Two & 76.87/73.62 & 70.82/67.74 & 78.07/77.52 & 70.09/69.58 & 81.31/73.27 & 73.79/66.18 & 71.22/70.07 & 68.57/67.47  \\
Pillar R-CNN-18 (12 epoch) & Two & 76.83/73.59 & 70.74/67.67 & 77.96/77.40 & 69.95/69.43 & 81.17/73.14 & 73.53/65.95 & 71.37/70.22 & 68.73/67.63 \\ 
\bottomrule       
\end{tabular}
}
\caption{Single-frame LiDAR-only non-ensemble 3D AP/APH performance comparison on the Waymo validation set.
The table is sorted by ALL APH/L2, which is the official leaderboard ranking metric.
 The proposed Pillar R-CNN demonstrates superiority over the state-of-the-art one-stage and two-stage approaches. $\dagger$: reported by LiDAR R-CNN \cite{li2021lidar}. }
\label{tab:waymo_val}
\end{table*}

\subsection{Comparison with State-of-the-Arts}

To validate the effectiveness of our proposed Pillar R-CNN, we fairly compare with state-of-the-art methods on the Waymo Open Dataset.

\vspace{-3mm}
\paragraph{Evaluation on the Waymo validation set.}
We compare our Pillar R-CNN with all published single-frame LiDAR-only non-ensemble methods on the Waymo validation set.
As shown in \cref{tab:waymo_val}, Pillar R-CNN achieves competitive performance for vehicle and pedestrian detection. 
By introducing FPN, our Pillar R-CNN achieves remarkably better AP/APH on all difficulty levels for the detection of small objects.
To be specific, our Pillar R-CNN with heavy PillarNet-34 backbone achieves 68.05 APH/L2 for pedestrian detection, surpassing the state-of-the-art AFDetV2 by +1.1\%.
The cyclist accuracy is hampered by the unbalanced object distribution as in \cite{shi2022pillarnet}, which dynamic label assignment techniques such as SimOTA \cite{yolox2021} can solve.
It is noteworthy that our Pillar R-CNN confirms the BEV representation can supply sufficient 3D structual information for superior detection accuracy. 

\vspace{-3mm}
\paragraph{Evaluation on the Waymo test set.}
We evaluate the performance of our Pillar R-CNN with its model variants on the Waymo Open Dataset \textit{test} set.
As shown in \cref{tab:waymo_test}, our Pillar R-CNN-34 outperforms all the previous single-frame LiDAR-only non-ensemble detectors, especially for vehicle and pedestrian detection. Our Pillar R-CNN-18/34 consistently surpass their one-stage counterparts PillarNet-18/34.
It is worth noting that the reported results of Pillar R-CNN apply $4\times$ stridden pooling map for refining 3D proposals on all three classes.

\textbf{Better results of integrating latest advances of 2D detection field will be provided later.}

\subsection{Ablation Studies}

We investigate the individual components of our proposed Pillar R-CNN framework with extensive ablation experiments.
For efficiency, we uniformly sub-sample 25\% of the training sequences and evaluate the full validation sequences.

\begin{table}[h]
\small
\centering
\begin{tabular}{l|c|c|cc|cc}
\toprule
\multirow{2}{*}{Methods} & \multirow{2}{*}{FPN} & \multirow{2}{*}{Aux.} & \multicolumn{2}{c|}{Veh. (APH)} & \multicolumn{2}{c}{Ped. (APH)} \\
& & & L1 & L2 & L1 & L2 \\
\midrule
\multicolumn{5}{l}{\textit{PillarNet-18 Backbone}} \\
 \midrule
\multirow{2}{*}{RPN} & & & 74.55 & 66.84 & 64.75 & 57.85 \\
 & \checkmark & & 74.54 & 66.86 & 70.04 & 63.25 \\
\midrule
\multirow{4}{*}{R-CNN} & & & 76.32 & 68.53 & 70.11 & 62.26 \\
& & \checkmark & 76.55 & 68.64 & 70.21 & 62.46 \\
& \checkmark & & 76.22 & 68.17 & 72.27 & 65.16 \\
 & \checkmark & \checkmark & 76.30 & 68.40 & 72.47 & 65.38 \\
 \midrule
\multicolumn{5}{l}{\textit{PillarNet-34 Backbone}} \\
 \midrule
RPN & \checkmark &  & 75.64 & 67.93 & 70.76 & 64.11 \\
\midrule
R-CNN & \checkmark & \checkmark & 76.99 & 69.07 & 73.18 & 66.07 \\
\bottomrule
\end{tabular}
\caption{Effects of each component in our Pillar R-CNN framework on Waymo validation set.
}
\label{tab:ab_any}
\end{table}

\vspace{-3mm}
\paragraph{Effects of Pillar R-CNN components.} 
\cref{tab:ab_any} details how each proposed component influences the accuracy of our Pillar R-CNN on various pillar-based backbones.
Based on the PillarNet-18 backbone, the $1^{st}$ and $3^{rd}$ rows show that compared with the RPN, the RoI refinement stage increases significantly by +1.69\% APH/L2 for vehicle detection and +4.41\% for pedestrian detection.
The auxiliary supervision on grid points per RoI brings marginal gains shown in $4^{th}$ row while this branch is disabled in inference.
Moreover, introducing FPN into RPN in the $2^{nd}$ row achieves a large gain of +5.4\% APH/L2 on small object detection like pedestrian.
With the proposals computed by RPN with FPN, shown in the $6^{th}$ row, our Pillar R-CNN achieves a further gain of +2.13\% APH/L2 on pedestrian.
In addition, by using the heavy yet powerful PillarNet-34 backbone, the performance of Pillar R-CNN can be further boosted overall categories.

\begin{table}[t]
\small
\centering
\begin{tabular}{c|c|cc|cc}
\toprule
\multirow{2}{*}{Stride} & \multirow{2}{*}{LC} & \multicolumn{2}{c|}{Veh. (APH)} & \multicolumn{2}{c}{Ped. (APH)} \\
& & L1 & L2 & L1 & L2 \\
\midrule
8 & & 75.88 & 67.96 & 68.93 & 61.44 \\
8 & \checkmark & 75.96 & 68.08 & 68.97 & 61.48 \\
4 & & 75.77 & 67.89 & 68.85 & 61.32 \\
4 & \checkmark & 76.55 & 68.64 & 70.21 & 62.46 \\
2 & & 76.16 & 68.30 & 69.97 & 62.16 \\
2 & \checkmark & 76.49 & 68.55 & 70.82 & 63.13 \\
\bottomrule
\end{tabular}
\vspace{-2mm}
\caption{Effects of the used pooling map resolutions and lateral connection layer in our Pillar R-CNN framework on Waymo validation set.}
\label{tab:ab_lc}
\end{table}

\begin{table}[t]
\small
\centering
\begin{tabular}{c|cc|cc}
\toprule
\multirow{2}{*}{Grid size} & \multicolumn{2}{c|}{Veh. (APH)} & \multicolumn{2}{c}{Ped. (APH)} \\
& L1 & L2 & L1 & L2 \\
\midrule
$4 \times 4$ & 75.58 & 67.63 & 69.14 & 61.67 \\
$5 \times 5$ & 76.11 & 68.19 & 69.77 & 62.17 \\
$6 \times 6$ & 76.32 & 68.37 & 70.11 & 62.40 \\
$7 \times 7$ & 76.55 & 68.64 & 70.21 & 62.46 \\
$8 \times 8$ & 76.90 & 68.96 & 70.29 & 62.69 \\
$9 \times 9$ & 76.70 & 68.78 & 70.11 & 62.47 \\
\bottomrule
\end{tabular}
\caption{Effects of the different grid sizes in our proposed RoI-grid pooling module on Waymo validation set.}
\label{tab:ab_grid}
\end{table}

\vspace{-5mm}
\paragraph{Effects of lateral connection layer.}
In \cref{tab:ab_lc}, we investigate the effects of the lateral connection layer used to build the pooling map. 
The $1^{st}$, $3^{rd}$ and $5^{th}$ rows of \cref{tab:ab_lc} show that pooling at fine resolutions alone does not improve performance.
It is because the upsampled top-down map has rich semantics but lacks precise spatial information.
The $2^{nd}$, $4^{th}$ and $6^{th}$ rows of \cref{tab:ab_lc} reveal that the designed lateral connection layer can encode high-level semantic and low-level geometry at various scales for boosting 3D detection performance.
Though simple, the lateral connection layer provides good semantic coverage on the sparse bottom-up map with mostly zeros.

\vspace{-5mm}
\paragraph{Effects of pooling map resolution.}
\cref{tab:ab_lc} illustrates the effects of the pooling map resolution with different spatial sizes constructed by our lateral connection module. 
The $2^{nd}$, $4^{th}$ and $6^{th}$ rows of \cref{tab:ab_lc} show that the finer resolution of the pooling map achieves better performance, especially for small objects.
However, a finer pooling map takes more additional memory/computation costs. 
Therefore, we use the $4 \times$ stridden pooling map as a trade-off throughout experiments, because $2 \times$ or finer pooling maps consume too much memory with marginal benefits.

\vspace{-3mm}
\paragraph{Effects of grid size per RoI.}
\cref{tab:ab_grid} shows the impact of different grid sizes within RoI-grid pooling module on detection performance. We can see that the accuracy consistently increases as the grid sizes from $4 \times 4$ to $8 \times 8$, but a larger grid size degrades performance slightly. 
The reason can be explained by that R-CNN with larger grid sizes has more learnable parameters in the first fully-connected layer and thus easily over-fits the training set.
Here, we adopt the grid size of $7 \times 7$ to keep the same setting with its 2D counterparts \cite{girshick2015fast,he2017mask}.

\begin{table}[t]
\small
\centering
\begin{tabular}{cccc|cc|cc}
\toprule
\multicolumn{4}{c|}{Bottom-up features} & \multicolumn{2}{c|}{Veh. (APH)} & \multicolumn{2}{c}{Ped. (APH)} \\
$C_1$ & $C_2$ & $C_3$ & $C_4$ & L1 & L2 & L1 & L2 \\ 
\midrule
& & \checkmark & & 76.55 & 68.64 & 70.21 & 62.46 \\
& & \checkmark & \checkmark & 76.34 & 68.45 & 70.15 & 62.39  \\
& \checkmark & \checkmark & & 76.75 & 68.82 & 70.41 & 62.65 \\
\checkmark & \checkmark & \checkmark & & 76.48 & 68.54 & 70.36 & 62.60 \\
\bottomrule
\end{tabular}
\caption{Effects of additional bottom-up features for constructing $4\times$ stridden pooling map on Waymo validation set.}
\label{tab:ab_mult}
\end{table}

\vspace{-3mm}
\paragraph{Effects of additional bottom-up features for pooling.}
Through our designed lateral connection, the built pooling map can integrate additional multi-scale features from the bottom-up pathway. 
The $2^{nd}$ row of \cref{tab:ab_mult} shows that combining coarser $C_4$ map slightly drops detection accuracy.
While the use of lower-level features in the $3^{rd}$ and $4^{th}$ rows yields negligible performance gains at an extra cost.
Hence, solely using its corresponding $4\times$ stridden bottom-up map achieves satisfying accuracy.

\subsection{Different Training Schemes}
\label{sec:regime}

We compare different two-stage training schemes, \textit{i.e.}, end-to-end training schedule and separate training of the RPN and R-CNN. 
Our Pillar R-CNN adopts the typical end-to-end training schedule for 36 epochs as \cite{shi2020pv,lin2017feature}.
This differs from our used codebase \cite{yin2021center}, where the RPN and R-CNN are separately trained with 36 epochs and 6 epochs respectively. 
We provide more experimental results of the separate training scheme, in which the RPN is first trained to generate region proposals and then frozen to train the R-CNN module. 
As shown in \cref{tab:waymo_val}, the 6-epoch schedule of Pillar R-CNN-18 results in slightly degraded detection performance, and a longer 12-epoch schedule may trap the R-CNN model in over-fitting.
In contrast, for Pillar R-CNN, it is easy to build a unified network in which the RPN and R-CNN are trained at once with favorable performance.

\begin{table}[th]
\small
\centering
\resizebox{0.99\linewidth}{!}{%
\begin{tabular}{l|c|c|cc|cc}
\toprule
\multirow{2}{*}{Methods} & \multirow{2}{*}{FPN} & \multirow{2}{*}{IoU} & \multicolumn{2}{c|}{Veh. (APH)} & \multicolumn{2}{c}{Ped. (APH)} \\
& & & L1 & L2 & L1 & L2 \\
\midrule
\multicolumn{5}{l}{\textit{PillarNet-18 Backbone}} \\
 \midrule
\multirow{2}{*}{RPN} & \checkmark & & 74.54 & 66.86 & 70.04 & 63.25 \\
 & \checkmark & \checkmark & 76.40 & 68.22 & 71.85 & 64.54 \\
\midrule
\multirow{2}{*}{R-CNN} & & & 76.55 & 68.64 & 70.21 & 62.46 \\
 & \checkmark &  & 76.30 & 68.40 & 72.47 & 65.38 \\
 \midrule
\multicolumn{5}{l}{\textit{PillarNet-34 Backbone}} \\
 \midrule
\multirow{2}{*}{RPN} & \checkmark &  & 75.64 & 67.93 & 70.76 & 64.11 \\
 & \checkmark & \checkmark & 77.30 & 69.16 & 73.38 & 66.23 \\
\midrule
R-CNN & \checkmark &  & 76.99 & 69.07 & 73.18 & 66.07 \\
\bottomrule
\end{tabular}
}
\caption{Effects of IoU-aware confidence rectification for our pyramidal region proposal network.}
\label{tab:iou}
\end{table}

\subsection{IoU-aware Confidence Rectification}
\label{sec:iou}

The IoU-aware confidence rectification is commonly used to cope with the misalignment issue for 2D detection \cite{jiang2018acquisition,wu2020iou} and 3D detection \cite{zheng2021cia,Hu2021AFDetV2RT} between localization accuracy and classification score.
The IoU-aware rectification function \cite{Hu2021AFDetV2RT,shi2022pillarnet} at the post-processing stage can be formulated as:
\begin{equation}
\centering
\hat{S} = S^{1-\beta} * W_{\rm{IoU}}^{\beta}
\end{equation}
where $S$ indicates the classification score and $W_{\rm{IoU}}$ is the IoU score. $\beta$ is a hyper-parameter.

To verify its effect on our proposed pyramidal region proposal network, we follow \cite{shi2022pillarnet} and conduct experiments for single-stage object detection.
From the results of RPN in \cref{tab:iou}, we can see that IoU-aware confidence rectification can achieve huge performance improvement on our pyramidal region proposal network.
After restoring using carefully tuned rectification factor $\beta$ in \cite{shi2022pillarnet}, the RPN achieves comparable performance with its R-CNN counterparts.
That is, the IoU-aware confidence rectification module benefits the ranking-based metrics such as Average Precision. 
In contrast, the R-CNN module mainly rescores boxes adaptively such that the boxes with better localization can be selected.

\section{Conclusion}
This paper presents a conceptually simple Faster R-CNN-like 3D detector, named Pillar R-CNN, for accurate 3D object detection solely on pillar-based point cloud representation.
The key insight lies in the pillar-based representation can offer crucial 3D structural information for accurate 3D detection. It differs from previous point-voxel-based two-stage methods that require abstracted intermediary keypoints for precise 3D box refinement. 
From this standpoint, we introduce FPN into RPN for better proposal generation on small objects and design a simple yet effective lateral connection layer to merge the low-level sparse pillar volumes and high-level dense semantic maps for the second 3D RoI refinement.
Moreover, Our Pillar R-CNN also takes into account the domain gap between image and point clouds for object detection, and attempts to design a novel two-stage detection paradigm to bridge this gap. 
As a result, our Pillar R-CNN builds a promising pathway for incorporating the advances from the well-developed 2D detection domain for accurate 3D detection based on pillar-based feature representation.
Extensive experimental results on the large-scale Waymo Open Dataset demonstrate that BEV representation on point clouds can preserve sufficient 3D structural information to facilitate the research of BEV perception.



{\small
\bibliographystyle{ieee_fullname}
\bibliography{egbib}
}

\end{document}